\documentclass[sigconf]{acmart}

\usepackage{censor}
\usepackage{todonotes}
\usepackage{url}
\usepackage{booktabs}
\usepackage{multirow}
\usepackage{tabularx}
\usepackage[acronym]{glossaries}

\AtBeginDocument{%
  \providecommand\BibTeX{{%
    \normalfont B\kern-0.5em{\scshape i\kern-0.25em b}\kern-0.8em\TeX}}}

\setcopyright{none}
\renewcommand\footnotetextcopyrightpermission[1]{} 
\begin{document}

\newacronym{LSTM}{LSTM}{Long Short Term Memory}
\newacronym{RNN}{RNN}{Recurrent Neural Network}
\newacronym{GRU}{GRU}{Gated Recurrent Unit}
\newacronym{CNN}{CNN}{Convolutional Neural Network}
\newacronym{ASJC}{ASJC}{All Science Journal Classification}
\newacronym{MLP}{MLP}{Multilayer Perceptron}
\newacronym{NLP}{NLP}{Natural Language Processing}
\newacronym{OOV}{OOV}{Out of vocabualry}

\title{Sequence-Based Extractive Summarisation for Scientific Articles}

\author{Daniel Kershaw}
\email{d.kershaw@elsevier.com}
\orcid{0000-0002-9130-5517}
\affiliation{%
  \institution{Elsevier Ltd}
  \streetaddress{125 London Wall}
  \city{London}
  \country{United Kingdom}
  \postcode{EC2Y 5AS}
}

\author{Rob Koeling}
\email{r.koeling@elsevier.com}
\orcid{0000-0002-8073-2409}
\affiliation{%
  \institution{Elsevier Ltd}
  \streetaddress{125 London Wall}
  \city{London}
  \country{United Kingdom}
  \postcode{EC2Y 5AS}
}

\begin{abstract}
This paper presents the results of research on supervised extractive text summarisation for scientific articles. We show that a simple sequential tagging model based only on the text within a document achieves high results against a simple classification model. Improvements can be achieved through additional sentence-level features, though these were minimal. Through further analysis, we show the potential of the sequential model relying on the structure of the document depending on the academic discipline which the document is from.
\end{abstract}



\maketitle

\section{Introduction}\label{sec:introductions}

The rate at which scientific knowledge is created and published is growing continuously~\cite{larsen2010rate}. Researchers are spending an increasing amount of time reading and understanding scientific documents. Automated summarisation is seen as a way of serving a more condensed version of an article to a reader in order to aid their reading experience and save time to acquire the necessary knowledge in their field. 

Research in automatic summarisation can roughly be separated in abstractive and extractive methods. Traditionally, extractive methods focused on core methods such as unsupervised graph-based methods \cite{GunesErkan:2005vn,W04-3252} and supervised model-based approaches~\cite{Collins:2017eua}.  Lately, there have been advances in the use of deep-learning for extractive and abstractive summarisation which resulted in the ability to get close to human quality automated summarisation~\cite{Nallapati:2017uq}.

This paper shows the results for applying sequential deep learning models to extractive summarisation on a large body of scientific publications. The models are trained and evaluated using author provided document summaries, consisting of a small number of take-away points highlighting the document's contributions.


Our main contributions are: 
\begin{enumerate}
    \item We introduce the use of a \acrshort{RNN} sequential tagger approach for extractive summarisation of scientific documents, which surpasses the baselines through introduction of global context
    \item The addition of hand-engineered features derived from sentences, improving the model, though not significantly
    \item We show that the importance of document structure differs between scientific domains
\end{enumerate}

We show the ability to apply extractive summarisation at scale for scientific documents, where the results are comparable to human performance. The  paper is structured as follows: related work in section~\ref{sec:relatedwork}, data sets are introduced in sections~\ref{sec:dataset} with the overarching model explained in section~\ref{sec:model}. Training of the model along with all model settings are discussed in section~\ref{sec:experiments}, followed by the results in section~\ref{sec:results} (this includes human evaluation). Finally, section~\ref{sec:discussion} concludes the work with a discussion of the results in the wider context. 

\section{Related Work}\label{sec:relatedwork}

Over the years, we have seen a wide variety of NLP research focussed on scientific articles: citations classification~\cite{Cohan:2019vy}, knowledge graph extraction~\cite{Luan:2018vn}, methods identification~\cite{Luan:2017wj} and citation networks~\cite{Stewart:2017vj} to name a few. Whereas some of these tasks are easy to explore with large open data-sets, others, such as knowledge graph creation, require access to curated annotated data; like those released for the SemEval 2017 and 2018 shared tasks~\cite{Augenstein:2017hb}.

The field of text summarisation has been furthered through the availability of corpora such as the CNN/DailyMail~\cite{See:2017ke}, or social media data-sets such as Reddit~\cite{Singer:2014kz}. This has lead to developments in abstractive and extractive summarisation models, such as a bidirectional~\acrshort{RNN} as the base of the model~\cite{Nallapati:2017uq}.      \citeauthor{Kedzie:2018wn} compared several extractive \acrshort{RNN} architectures and showed there were limited performance improvements through variations of the model architectures, such as the inclusion of attention~\cite{Narayan:2017ux}.

Focusing on the scientific domain~\citeauthor{Collins:2017eua} showed that one could use extractive summarisation on scientific articles by classifying each sentence as `summary like' or not. But as each sentence was independently classified, there were issues with the lack of global context. An alternative approach was proposed by ~\citeauthor{jaidka2019cl}, who used `Citations'\footnote{This is the contextual sentence around a citation} to generate a summary of the document based on what other documents have said about it. This approach results in summaries of what peers actually think of the research, rather than the authors themselves. Though traditionally, methods for summarising scientific articles focus on the patterns within the text with, for example,~\citeauthor{teufel-moens-2002-articles} looking at the rhetoric patterns within the text to extract summaries.

In this paper we show that through the application of neural extractive summarisation we can generate short (4 sentence), human readable summaries which match a human baseline on a large corpus of scientific documents from across a variety of scientific disciplines.

\section{Data Set}\label{sec:dataset}

In this research study we leverage so-called author-provided highlights to train a extractive summarization model. \textit{`Author highlights'} consist of four to six bullet points provided by the author at time of submission, and are designed to be a \textit{`key finding'} summary of the paper, which is more condensed than the abstract. Author highlights are similar to the summaries provided within the CNN/DailyMail data-set~\cite{See:2017ke}, which is used extensively in summarisation research. Example author highlights are found in Table~\ref{tab:example_highlights}.

The data-set contains $138,735$ documents, this is split into train, test and validation sets. The train and test sets are used during the model training and the validation set is used to report the results presented in sections~\ref{sec:results}. The documents in this data-set have on average $4$ author highlight, with an average of $12$ tokens per highlight. This is in comparison to the CNN/DailyMail data-sets which has $4$ summary sentences, with a average length of $13$ tokens. One notable difference between the two data-sets is that while the CNN/DailyMail documents have on average $22$ and $30$ sentences respectively, the scientific documents have an average of $128$ sentences. Other notable differences are: the more complex language and a higher number of \acrshort{OOV} terms used in scientific writing and the different structure in which the articles are written. 




\subsection{Data Sampling}\label{subsec:sampeling}

\begin{table*}[t]
\centering
\begin{tabular}{lllll}
\hline
              & \textbf{\# Documents}               & \textbf{Average Labels}   & \textbf{Average \# Sentence} & \textbf{Avg sentence length  }     \\ \hline
Test          & \multicolumn{1}{r}{41,756} & \multicolumn{1}{r}{10.04} & \multicolumn{1}{r}{162.85} & \multicolumn{1}{r}{24.07} \\
Train         & \multicolumn{1}{r}{83,153} & \multicolumn{1}{r}{10.02} & \multicolumn{1}{r}{163.62} & \multicolumn{1}{r}{24.06}  \\
Validation & \multicolumn{1}{r}{13,826} & \multicolumn{1}{r}{10.07} & \multicolumn{1}{r}{163.51} & \multicolumn{1}{r}{24.13} \\ \hline
\end{tabular}
\label{tab:general_stats}
\caption{General statistics for the data-sets depending on the domains they represent}
\end{table*}

The training data for the model is generated by selecting sentences from within the documents that are `similar to' the author provided highlights. \citeauthor{Nallapati:2017uq} proposed the use of a greedy sampling method to select the best sub-set of `similar' sentences. In greedy sampling one sentence is added to the selected set at a time, based on which one increases the overall similarity metric the most. For this work we use \texttt{rouge-l-f} as the metric to compare sentences to the set of highlights. Selection of new sentences is stopped once $10$ sentences have been selected. Unlike~\citeauthor{Kedzie:2018wn} and~\citeauthor{Collins:2017eua}, who optimised for \texttt{rouge-2-r}, \texttt{rouge-l-f} was chosen as it was shown to yield balanced summaries which contain the core topics. This can be seen as a trade-off between recall based metrics, which tend to produce large amounts of noise, and precision based metrics that favour the shortest summaries.

The articles within our data contain a number of features not seen in other similar data-sets. Scientific articles are written in a more formalised structure, with the documents broken down into distinct (sub)sections to aid the reader in understanding the content. Using a gazetteer which maps section titles to a set of high level section types\footnote{Introduction, Methods, Results, Discussions, Conclusion, Other}, it was shown that $29.9\%$ of sentences were taken from the `Results' section. This would be in line with the author highlights focusing on the results and findings of the article. 

Additionally, author highlights are on average $12$ tokens long, compared to an average of $15$ tokens for the sentences sampled from the document using \texttt{rouge-l-f}. Sampling with \texttt{rouge-l-r} results in much longer sentences: an average of $28$ tokens. This is evidence that the sentences sampled using \texttt{rouge-l-f} are more similar to the author provided highlights. 

\subsection{Preprocessing} \label{subsec:preprocessing}

Before the documents are used to train the model they go through a number of preprocessing steps. We normalise or remove non-standard elements typical for scientific documents, like mathematical formulae, chemical compounds and maps. In addition custom XML mark-up is removed, which could contain citation information. This leaves us with the plain text, which is normalised and tokenized using the \acrshort{NLP} processing pipeline from JohnSnow Lab.\footnote{\url{https://nlp.johnsnowlabs.com/}}

\section{Model} \label{sec:model}

We are aiming to select the best subset of sentences,  representing the main take-away points (highlights) from an article. As with~\cite{Kedzie:2018wn} and~\cite{Nallapati:2017uq} we treat this as a sequence tagging problem. Given an article $D$ that contains a set of $n$ ordered sentences $s_0, s_1, ...  s_n$, a summary is generated by predicting the label of the sentences $y_1, ..., y_n \in \{0,1\}^n$, such that $y_i = 1$ means the $i^{th}$ sentence should be included in the summary. 

At a high level the proposed models  can be broken down into two components, the sentence encoder (see section~\ref{subsec:sentence_encoder}) and the sentence extractor (see section~\ref{subsec:sentenceextractor}). The sentence encoder processes each individual sentence to form a vector representation, and the sentence extractor then takes all sentence encodings in order to select the best sentences to be included in the  highlights. This means that given a sequence of sentence embeddings $h = h_1, h_2, ..., h_n$ the extractor outputs a sequence of predictions (probabilities) in the form of $y = y_1, y_2, ..., y_n$. 

\citeauthor{Kedzie:2018wn}~showed there is limited difference between the results of varying \acrshort{RNN} sequence tagging extractive summarisation models, such as~\cite{cheng-lapata-2016-neural} which included an attention mechanism, or~\citeauthor{Nallapati:2017uq} who processed the document twice.
For this reason a simple single layered bi-direction \acrshort{RNN} was chosen as the base of the summarisation model where the classification was made on the concatenation of the forward and backwards hidden states of each sentence which is then passed through a~\acrfull{MLP}. 

\subsection{Sentence Encoders}\label{subsec:sentence_encoder}

\citeauthor{Kedzie:2018wn} proposed three distinct sentence encoders, each representing the sentences in different ways. These are: averaging word embeddings (MEAN), \acrshort{RNN} and \acrshort{CNN} encoding of the word vectors.

\paragraph{MEAN}
For a sentence of $n$ words, the embedding is simply the average over the set of word embeddings $h = \frac{1}{|s|}\sum_{1=1}^{|s|}$

\paragraph{\acrshort{CNN}} A number of $1D$ \acrshort{CNN}'s with varying widths are passed over the word embeddings to produce a sentence embedding. This is similar to~\cite{Anonymous:uhC_Oukc}, though instead of classifying the final layer with a number of dense layers, it is used as an input for the next level of the model.

\paragraph{\acrshort{RNN}} As in~\cite{Collins:2017eua} sentences can be represented as the concatenated final states of a single layered bi-directional \acrshort{RNN}. Though instead of using a \acrshort{GRU} for the \acrshort{RNN} cell an \acrshort{LSTM}~\cite{gers1999learning} is used, to replicate the sentence encoding found in~\cite{Collins:2017eua}.

\subsubsection{Additional Features}\label{subsec:sentence_modification}

\citeauthor{Collins:2017eua} and~\citeauthor{Narayan:2017ux} demonstrated that both extractive and abstractive summarisation models can be improved with the addition of pre-computed features (side information) in the model. This is achieved by concatenating additional features on to the sentence encodings. For this work, the sentence level features proposed in~\cite{Collins:2017eua} are used. These are:

\noindent \textbf{Number of numbers} - the number of numeric tokens within the sentence.

\noindent \textbf{Sentence Length} - the number of tokens within the sentence.

\noindent \textbf{Section Classification} - a one-hot encoding of the section class where the sentence comes from. This can either be `Introduction', `Related Work', `Methods', `Results', `Discussions', `Conclusion' or `Other'. This is done through a simple string match against a gazetteer of section titles.

\noindent \textbf{Title Overlap} - the normalised number of words which appear in both the sentence and the title.

\noindent \textbf{Key Phrase Overlap} - the number of words which appear in both the key term list and sentence.

\noindent \textbf{Abstract Overlap} - the number of words which appear in both the sentence and abstract of the document.

These features are concatenated together into one feature vector per sentence. The feature vector is then passed through a number of dense layers before being concatenated with the sentence embedding. This then makes up the input for the document encoder.

\subsection{Sentence Extractors}\label{subsec:sentenceextractor}

\citeauthor{Kedzie:2018wn} compared a number of sequence based models for extractive summarisation. For simplicity this paper focuses on a simple single layered bi-directional \acrshort{RNN} based sequence tagging model. The output of the forward and backwards pass of each cell within the \acrshort{RNN} are concatenated and then passed to a \acrshort{MLP} layer for final classification. The \texttt{soft-max} output of the classifier is then regarded as the probability whether the sentence should be included in the summary or not. This is a discriminatory classifier $p(y_{1:n}|h_{1:n})$. This means that each prediction is made independent of the other predictions by the model. As with one of the sentence encoders proposed in section~\ref{subsec:sentence_encoder} \acrshort{LSTM} cells are used within the \acrshort{RNN}. 

\subsubsection{Modification}

Section~\ref{subsec:sentence_modification} introduced sentence level features in addition to the sentence embedding. At a document level there are also features which can be used within the sentence extractor. Though instead of concatenating them to an output of the model (as with sentences in section~\ref{subsec:sentence_modification}), the hidden states of the \acrshort{LSTM} cells within the \acrshort{RNN} are initialised with document level features. Initialising the hidden state with user and product attributes was done by~\citeauthor{Ni:2018tg,Ni:2017vr,Ni:2019fb} to generate personal abstractive reviews for products. Instead of using product and user based features, we initialise the \acrshort{LSTM} with the \acrshort{ASJC} codes, the abstract and the title for the given document. 

\noindent \textbf{ASJC} - each document is given a series of codes which represent the subject area and disciplines\footnote{There are $334$ unique \acrfull{ASJC} codes, each representing disciplines and sub-disciplines. More information on them can be found at \url{https://service.elsevier.com/app/answers/detail/a_id/15181/supporthub/scopus/}} of the journal in which the paper was published. A journal can be associated with multiple \acrfull{ASJC} codes, thus for a document the final \acrshort{ASJC} vector is the normalised sum of the individual \acrshort{ASJC} embeddings. 

\noindent \textbf{Title} - the title is represented by the average of its word embeddings 

\noindent \textbf{Abstract} - the abstract is represented by the average of its word embeddings

The vectors which represent these three additional features are concatenated together and passed though a fully connected layer before being used to initialise the \acrshort{LSTM} within the sentence extractor. 

\subsection{Base Model}\label{subsec:baseline}

We approach extractive summarisation as a sequence tagging problem. In contrast, \citet{Collins:2017eua} classified each individual sentence independently. Thus, where in the sequence tagging model the preceding sentences influence the prediction for the current sentence, there is no interaction between sentences in the classification model. We use the latter model as a baseline in the evaluation stage. 

\section{Experiments}\label{sec:experiments}

To evaluate the quality of the automated summaries they are compared to the gold standard (author highlights) provided by the authors. The comparison is done on the top 4 ranked sentences, using \texttt{rouge-l-f}. The metric from here on will be referred to as \texttt{rouge-l-f@4}. 

\subsection{Settings}

We initialise the model with \texttt{GLOVE} embeddings of size `100'. Unknown words receive an embedding which has been randomly initialised. Different models were trained with and without the ability to modify the embeddings (see results in Table~\ref{tabel:results_sentence_encoder}).

\paragraph{Sentence Encoder} First the three proposed sentence encoders are tested. A number of the parameters are held constant due to results from previous experimentation. The word embedding dimensionality is $100$ across all the experiments and all \acrshort{LSTM} cells within the sentence encoder have a hidden state of $100$. For the \acrshort{CNN} we use $25$ filters of size  $[1,2,3,4]$. This means all three sentence encoders produce output vectors of size $100$. 

\paragraph{Sentence Extractor} The hidden state within the \acrshort{LSTM} cell is consistently held at $128$. The \acrshort{MLP} layer is of size $50$ before it is classified using a softmax layer. The \acrshort{ASJC} embedding size is set to a size of $100$ which is used in the additional features along with the title and abstract embeddings (which have a length of $100$)

\subsection{Training}

Models are trained using weighted negative log-likelihood which has to be minimised. 

\begin{equation} \label{eq:neg-log-loss}
    L = - \sum_{s,y \in D} \sum^{n}_{i=1}  \omega (y_i) \log p (y_i | y_{<i}, h)
\end{equation}

The weights within the models are proportional to the number of positive and negative labels where $w(0) = 1$ and $w(1) = N_1/N_0$. Where $N_n$ is the number of labels within the document for that classification.

The model was optimised using \texttt{ADAM}~\cite{kingma2014adam} and stochastic gradient descent, with a learning rate of $0.0001$ and dropout across the model of $0.25$. Additionally, gradient clipping was used to mitigate against the problem of vanishing gradient, this was set to $1$. Models are trained for up-to $50$ epochs, with early stopping if there has been no decrease in validation \texttt{loss} for $5$ epochs. 

\section{Results}\label{sec:results}

The following section reports the results of the experiments outlined above. First the results for the varying sentence encoders are reported, with the best performing sentence encoder used to test the effectiveness of adding additional features to the model. The best performing model overall is submitted for human evaluation in sections~\ref{subsec:human_evaluation}. 

All models are trained three times, and the results reported are created using a document level average across each of the runs. Additionally, an approximate randomisation test is used to report on statistic confidence in the results between models.\footnote{\url{https://github.com/Sleemanmunk/approximate-randomization}}

\subsection{Sentence Encoder}\label{subsec:sentence_encoder_experiment}

\begin{table*}[t]
\centering
\begin{tabular}{llrrrr}
\hline
                     &          & \textbf{Biology} & \textbf{Computing} & \textbf{Economics} & \textbf{All}   \\ \hline
\multirow{2}{*}{CNN} & False(*) & 20.87            & \textbf{23.01}     & \textbf{21.92}     & 21.91          \\
                     & True     & \textbf{21.03}   & 22.97              & 21.61              & \textbf{22.19} \\
\multirow{2}{*}{MAN} & False(*) & 20.75            & 22.50              & 20.99              & 21.79          \\
                     & True(*)  & 20.90            & 22.44              & 21.02              & 21.89          \\
\multirow{2}{*}{RNN} & False(*) & 20.85            & 22.60              & 21.31              & 21.86          \\
                     & True(*)  & 20.49            & 22.33              & 20.97              & 21.51          \\ \hline
\end{tabular}
\caption{\texttt{rouge-l-f@4} scores for each of the six variations in sentence encoders. Models with a * indicate they are statistically worst than the best performing model ($p > 0.05$)}
\label{tabel:results_sentence_encoder}
\end{table*}

Results for the varying sentence encoders can be found in Table~\ref{tabel:results_sentence_encoder}. The metrics reported are \texttt{rouge-l-f@4} against the author provided highlights in the validation set. Results indicate that the best performing models are those developed with the \acrshort{CNN} based sentence encoder. This is consistent with the results when they are broken down by discipline. As one can see the results show limited variance between the different forms of sentence encoders. Moreover, within them all offer the same stability across disciplines.  The same stable scores across the sentence encoders can be seen in~\cite{Kedzie:2018wn}, where no significant difference was reported between the models. This resulted in the MEAN embeddings being used. However, here the \acrshort{CNN} based embedding with trainable word embeddings is used in the next experiment. 

\subsection{Additional Features}\label{subsec:additinal_featrures_experiment}

\begin{table*}[t]
\centering
\begin{tabular}{llrrrr}
\hline
\textbf{Sentence Features}    & \textbf{Document Feature}   & \textbf{Biology}     & \textbf{Computing}   & \textbf{Economics}   & \textbf{All}         \\ \hline
False                         & False                       & 21.03 (0.00)                & 22.97 (0.00)                & 21.61 (0.00)                & 22.19 (0.00)                \\
                              & True                        & 20.88 (-0.15)                & 23.34 (0.35)               & 21.49 (-0.12)                & 22.14 (-0.05)               \\
\textbf{True}                 & \textbf{False}              & \textbf{21.45} (0.42)       & \textbf{23.42} (0.45)      & \textbf{21.97} (0.36)      & \textbf{22.37} (0.18)      \\
                              & True(*)                     & 20.57 (-0.46)               & 22.64 (-0.33)               & 21.05 (-0.56)                & 21.59 (-0.6)                \\ \hline
\textbf{Baseline}                      &                             & \multicolumn{1}{l}{} & \multicolumn{1}{l}{} & \multicolumn{1}{l}{} & \multicolumn{1}{l}{} \\ \hline
\citet{Collins:2017eua} &                            & 13.12                                & 13.42              & 16.61              & 12.96            \\ \hline
\end{tabular}
\caption{\texttt{rouge-l-f@4} scores for the \acrshort{CNN} based sentence encoder model with additional features on both the sentence and document level. Models marked with a * when compared to the best model are statistically worse than the best performing model ($p > 0.05$). Value in brackets indicate difference between best performing model in Table~\ref{tabel:results_sentence_encoder}}
\label{tabel:results_additinal_features}
\end{table*}

As proposed in sections~\ref{subsec:sentence_modification} and~\ref{subsec:additinal_featrures_experiment} the model was modified with the addition of  features to both the sentence encoder and sentence extractor. In Table~\ref{tabel:results_additinal_features} we show the results for the \acrshort{CNN} based sentence encoder with trainable embeddings with the additional features. The inclusion of additional features with the sentence encoder results in a noticeable improvement in the \texttt{rouge-l-f@4} score, with computing articles seeing a maximum increase of $0.45$. However, even though the modification of the sentence encoder improves the quality of the summaries, initialising the sentence extractor with the document level features degrades the quality of the summaries slightly. The document features on their own reduced the \texttt{rouge-l-f@4} by $0.05$, though not statistically significant. 

\subsection{Baseline}

The initial model discussed in section~\ref{subsec:sentenceextractor} with the addition of document and sentence level features significantly outperform the baseline model proposed in~\cite{Collins:2017eua}. The baseline model classified each sentence separately from each other, meaning there is no addition of context in the classification from the surrounding sentences. The significant difference in the results from the two models indicates that the inclusion of more contextual information through the \acrshort{RNN} is important. This is evidence that structure is an intrinsic feature within the model.

\subsection{Structural Analysis} \label{subsec:further_analysis}

The training data-set is constructed using greedy sampling, with the majority of sentences coming from the `Results' section of the articles. When looking at the distribution of the location of the predicted sentences within the document, one can see that there is indeed a dependency on the structure of the document on which sentences are selected. Within the top 4, $28.84\%$ of sentences are from the `Results' section, followed by sentences from the `Introduction' which accounted for $19.79\%$. This distribution of sentences from across the article is not seen in the summarisation of news articles. \cite{hong-nenkova-2014-improving} reported a strong lead-bias, where sentences from the beginning of the articles dominate the summaries.

\begin{table}[]
\centering
\small
\begin{tabular}{@{}lrrrr@{}}
\toprule
\textbf{Shuffled} & \textbf{Biology} & \textbf{Computing} & \textbf{Economics} & \textbf{All} \\ \midrule
False              & 21.03            & 22.97              & 21.61              & 22.19        \\
True               & 20.11               &  22.62                 &  21.09                & 20.75       \\ \bottomrule
\end{tabular}
\caption{\texttt{rouge-l-f@4} scores for the \acrshort{CNN} based sentence encoder model without additional features.}
\label{tabel:results_shuffeled}
\end{table}

To further investigate how much the model depends on the structure of the document, we train the model again with the sentences within each training document shuffled. The new model is then applied to the validation data-set which has not been shuffled. A significant drop in the metrics would then indicate that the structure of the document and the context which is learnt through the document level~\acrshort{RNN} is important.

Results can be seen in Table~\ref{tabel:results_shuffeled}. As one can see there is a drop in performance across all disciplines, with the \texttt{rouge-l-f@4} reducing from from $22.19$ to $20.75$, a $9.35\%$ reduction. The significant reduction across the board is similar to results reported in~\cite{Kedzie:2018wn} for news, showing that the model is learning the position of the sentence within the document. This is understandable since scientific articles generally have a clear, intrinsic structure. Though when looking at the individual disciplines one can see that the size of the reduction varies across the board. Disciplines such as Economics and Computing, which are associated with a more varied writing style, have less of a reduction. This would indicate that for some disciplines the structure of documents is more important than for others. 

\subsection{Length Analysis}

Author based highlights had on average $11.92$ tokens, while we established that the sampled sentences (using \texttt{rouge-l-f}) are $15$ tokens long. The best performing model extracted sentences with an average of $11.36$ tokens. This is compared to the $32.43$ tokens reported by~\citeauthor{Collins:2017eua}. This shows that the model is not only looking at the text and location of the sentence but also the density of information within the sentence. Resulting in short highlights which are heuristically similar to the author provided highlights. 

\subsection{Human Evaluation} \label{subsec:human_evaluation}
\begin{table*}[t]
    \centering
    \small
    \begin{tabularx}{\textwidth}{p{0.45\textwidth}p{0.45\textwidth}}
    \toprule
    \textbf{Author Highlight} & \textbf{Automated Summary} \\ \midrule
   \begin{itemize}
   \itemsep0em 
\item The expression pattern of visfatin was ubiquitous in the various avian tissues. 
\item Visfatin mRNA was most highly expressed in breast muscle and continuously decreased with increasing age in silky fowl. 
\item Subcutaneous fat and visceral fat exhibited higher contents of visfatin mRNA in broiler chicken than those in silky fowl. 
\item Visfatin fusion protein significantly increased the expression of adipocyte differentiation marker genes.
\end{itemize}
                          &  
                          \begin{itemize}
                          \itemsep0em 
\item Both subcutaneous fat and visceral fat exhibited higher contents of visfatin mRNA in broiler chickens than those in silky fowl. 
\item 3T3-L1 adipocytes were treated with recombinant chicken visfatin and insulin. 
\item Furthermore, visfatin fusion protein significantly increased the expression of adipocyte differentiation marker genes. 
\item The visfatin mRNA levels continuously decreased with increasing age in silky fowl.
\end{itemize}                          \\ \bottomrule
    \end{tabularx}
\caption{Example author highlights and automated summary for the document ``Characterization of the visfatin gene and its expression pattern and effect on 3T3-L1 adipocyte differentiation in chickens'' (\url{https://www.sciencedirec.com/science/pii/S0378111917306765})}
\label{tab:example_highlights}
\end{table*}

Computed metrics for text summarisation such as \texttt{rouge} only give an indication about the quality of the automated summary, thus human judgement was included. $7$ human annotators were tasked with rating a set of $12$ automated summaries, created using the best performing model (\acrshort{CNN} sentence encoder, with additional sentence features), each document was assessed $3$ times. The raters were not necessarily experts in the domains of the documents they assessed. The articles included in the test were a random sample from across academic disciplines.

Raters were asked to rate each set of summaries on a scale of $1$ (low) to $4$ (high) on four dimensions:

\begin{itemize}
\itemsep0em 
\item \textbf{Simplicity}: are the sentences which have been selected simple to read or are they too long and using over-complicated language.
\item \textbf{Informativeness}: do the sentences which have been selected inform the user about what is going on within the papers
\item \textbf{Relevant}: are the sentences which have been selected relevant to the main findings of the paper
\item \textbf{Diversity}: are all the sentence which have been selected covering the same points or is there diversity across the sentences. 
\end{itemize}

In order to get a baseline, a number of author-generated summaries (gold standard) were included in the rating task. This allowed us to get a side by side comparison of how the best trained model performed compared to the gold standard author provided highlights. Inter-annotator agreement for the four dimensions was: $72.92$, $66.67$, $70.83$, $54.17$ respectively. 

\begin{table}[h]
\centering
\begin{tabular}{lrrrr}
\hline
\textbf{Discipline} & \multicolumn{1}{l}{\textbf{Dive}} & \textbf{Info} & \textbf{Simp} & \textbf{Rele} \\ \hline
Bio Science         & 2.81                              & 2.49          & 2.93          & 2.65          \\
Computing           & 2.95                              & 1.57          & 2.95          & 1.86          \\
Economic            & 3.00                              & 2.67          & 3.67          & 2.33          \\
All                 & 2.90                              & 2.37          & 2.83          & 2.51          \\ \hline
Author highlights       & 3.01                              & 2.78          & 2.81          & 3.2          \\ \hline
\end{tabular}
\caption{Results for human evaluation}
\end{table}

The raters assessed the author-provided summaries higher than the automated ones. However, the ratings for the automated highlights are not significantly worse than the author provided ones. The assessors judged that the automated summaries were both simple and diverse. It should be pointed out that there was limited inter-annotator agreement on the diversity, indicating potential misunderstanding of this dimension. The summaries were rated lower (but still favourably) on the informative and relevance scale.

\section{Discussion}\label{sec:discussion}

Extracting simple automated summaries from content is challenging. The research presented in this paper demonstrates the ability to successfully apply neural extractive text summarisation methods to large scientific articles from a variety of domains. This work shows that moving from a binary classification based model to one which is sequential in nature can result in a significant uplift in the performance of the model. This means the model not only learns from the text but also takes the structure of the scientific article into account, bringing in local and global context to each prediction. 

The positive results were observed across a variety of scientific domains (including social sciences). The robustness of the models across domains indicates that the model is potentially learning from common phrase structures seen across scientific disciplines. This would then explain why the inclusion of additional article and sentence-level features resulted in marginal increases in model performance. It would also explain the inclusion of sentences containing common phrases like `Results can be found in section' in several summaries. The dependency on the structure of the articles seems to differ across disciplines, with a subject like Biology being more impacted in the sentence shuffle experiment. It should also be noted that sentences selected from the methods and results sections often include \acrshort{OOV} words such as proteins and chemicals, thus potentially indicating that the sentence encoders were successfully picking up common patterns within the phrasing of the text.

Through the use of human evaluation, we showed that the automated highlights generated were comparable in quality to those submitted by the author. This could be interpreted two ways though, either the model is doing well, or the highlights submitted by the authors are not that good. Speaking to the raters and doing manual inspection of the summaries taught us that it could be a combination of the two. This can be seen in several examples where the author highlights in the validation set where taken directly from the article, with the model then predicting the correct sentences. 

From a technical perspective one could argue that even though across the board the \acrshort{CNN} achieves the best results, it might not the best model to deploy. A major drawback of this model is the length of time it takes to train compared to other models. Training the \acrshort{CNN} model takes on average $24$ hours, compared to $11$ hours for the MEAN model.\footnote{Trained on an AWS \texttt{ml.p3.16xlarge} instance} The addition of sentence and article-level features resulted in marginal improvements, which may not have justified their inclusion as they again take time to compute. 

\section{Conclusion}

To conclude, in this paper we presented an empirical analysis of using sequential models for extractive text summarisation. Results indicate that through the application of a basic \acrshort{RNN} model, one can get summaries which are as good as those provided by the author of the paper. Depending on the disciplines of the paper there are varying degrees of reliance on the structure of the article. In further research we will look at the use of more complex sentence embeddings, which can take into account tokens such as protein and genes names which are currently treated as random embeddings. 

\bibliographystyle{ACM-Reference-Format}
\bibliography{main}


\begin{thebibliography}{24}


\ifx \showCODEN    \undefined \def \showCODEN     #1{\unskip}     \fi
\ifx \showDOI      \undefined \def \showDOI       #1{#1}\fi
\ifx \showISBNx    \undefined \def \showISBNx     #1{\unskip}     \fi
\ifx \showISBNxiii \undefined \def \showISBNxiii  #1{\unskip}     \fi
\ifx \showISSN     \undefined \def \showISSN      #1{\unskip}     \fi
\ifx \showLCCN     \undefined \def \showLCCN      #1{\unskip}     \fi
\ifx \shownote     \undefined \def \shownote      #1{#1}          \fi
\ifx \showarticletitle \undefined \def \showarticletitle #1{#1}   \fi
\ifx \showURL      \undefined \def \showURL       {\relax}        \fi
\providecommand\bibfield[2]{#2}
\providecommand\bibinfo[2]{#2}
\providecommand\natexlab[1]{#1}
\providecommand\showeprint[2][]{arXiv:#2}

\bibitem[Sin(2014)]%
        {Singer:2014kz}
 \bibinfo{year}{2014}\natexlab{}.
\newblock \bibinfo{booktitle}{\emph{{Evolution of reddit}}}.
  \bibinfo{publisher}{ACM Press}, \bibinfo{address}{New York, New York, USA}.
\newblock
\showISBNx{9781450327459}
\urldef\tempurl%
\url{https://doi.org/10.1145/2567948.2576943}
\showDOI{\tempurl}


\bibitem[Augenstein et~al\mbox{.}(2017)]%
        {Augenstein:2017hb}
\bibfield{author}{\bibinfo{person}{Isabelle Augenstein},
  \bibinfo{person}{Mrinal Das}, \bibinfo{person}{Sebastian Riedel},
  \bibinfo{person}{Lakshmi Vikraman}, {and} \bibinfo{person}{Andrew McCallum}.}
  \bibinfo{year}{2017}\natexlab{}.
\newblock \showarticletitle{{SemEval 2017 Task 10 - ScienceIE - Extracting
  Keyphrases and Relations from Scientific Publications.}}
\newblock \bibinfo{journal}{\emph{SemEval@ACL}} (\bibinfo{year}{2017}),
  \bibinfo{pages}{546--555}.
\newblock
\urldef\tempurl%
\url{https://doi.org/10.18653/v1/S17-2091}
\showDOI{\tempurl}


\bibitem[Cheng and Lapata(2016)]%
        {cheng-lapata-2016-neural}
\bibfield{author}{\bibinfo{person}{Jianpeng Cheng} {and}
  \bibinfo{person}{Mirella Lapata}.} \bibinfo{year}{2016}\natexlab{}.
\newblock \showarticletitle{Neural Summarization by Extracting Sentences and
  Words}. In \bibinfo{booktitle}{\emph{Proceedings of the 54th Annual Meeting
  of the Association for Computational Linguistics (Volume 1: Long Papers)}}.
  \bibinfo{publisher}{Association for Computational Linguistics},
  \bibinfo{address}{Berlin, Germany}, \bibinfo{pages}{484--494}.
\newblock
\urldef\tempurl%
\url{https://doi.org/10.18653/v1/P16-1046}
\showDOI{\tempurl}


\bibitem[Cohan et~al\mbox{.}(2019)]%
        {Cohan:2019vy}
\bibfield{author}{\bibinfo{person}{Arman Cohan}, \bibinfo{person}{Waleed
  Ammar}, \bibinfo{person}{Madeleine van Zuylen}, {and} \bibinfo{person}{Field
  Cady}.} \bibinfo{year}{2019}\natexlab{}.
\newblock \showarticletitle{{Structural Scaffolds for Citation Intent
  Classification in Scientific Publications}}.
\newblock \bibinfo{journal}{\emph{arXiv.org}} (\bibinfo{date}{April}
  \bibinfo{year}{2019}).
\newblock
\showeprint[arxiv]{1904.01608v2}~[cs.CL]
\urldef\tempurl%
\url{http://arxiv.org/abs/1904.01608v2}
\showURL{%
\tempurl}


\bibitem[Collins et~al\mbox{.}(2017)]%
        {Collins:2017eua}
\bibfield{author}{\bibinfo{person}{Ed Collins}, \bibinfo{person}{Isabelle
  Augenstein}, {and} \bibinfo{person}{Sebastian Riedel}.}
  \bibinfo{year}{2017}\natexlab{}.
\newblock \showarticletitle{{A Supervised Approach to Extractive Summarisation
  of Scientific Papers.}}
\newblock \bibinfo{journal}{\emph{CoRR}} (\bibinfo{year}{2017}),
  \bibinfo{pages}{195--205}.
\newblock
\urldef\tempurl%
\url{https://doi.org/10.18653/v1/K17-1021}
\showDOI{\tempurl}


\bibitem[Erkan and Radev(2004)]%
        {GunesErkan:2005vn}
\bibfield{author}{\bibinfo{person}{G{\"u}nes Erkan} {and}
  \bibinfo{person}{Dragomir~R Radev}.} \bibinfo{year}{2004}\natexlab{}.
\newblock \showarticletitle{{LexRank - Graph-based Lexical Centrality as
  Salience in Text Summarization.}}
\newblock \bibinfo{journal}{\emph{Journal of Artificial Intelligence Research}}
   \bibinfo{volume}{22} (\bibinfo{year}{2004}), \bibinfo{pages}{457--479}.
\newblock
\urldef\tempurl%
\url{https://www.scopus.com/inward/record.uri?partnerID=HzOxMe3b&scp=27344433526&origin=inward}
\showURL{%
\tempurl}


\bibitem[Gers et~al\mbox{.}(2000)]%
        {gers1999learning}
\bibfield{author}{\bibinfo{person}{Felix~A Gers}, \bibinfo{person}{J{\"u}rgen
  Schmidhuber}, {and} \bibinfo{person}{Fred~A Cummins}.}
  \bibinfo{year}{2000}\natexlab{}.
\newblock \showarticletitle{{Learning to Forget - Continual Prediction with
  LSTM.}}
\newblock \bibinfo{journal}{\emph{Neural Computation}} \bibinfo{volume}{12},
  \bibinfo{number}{10} (\bibinfo{year}{2000}), \bibinfo{pages}{2451--2471}.
\newblock
\urldef\tempurl%
\url{https://doi.org/10.1162/089976600300015015}
\showDOI{\tempurl}


\bibitem[Hong and Nenkova(2014)]%
        {hong-nenkova-2014-improving}
\bibfield{author}{\bibinfo{person}{Kai Hong} {and} \bibinfo{person}{Ani
  Nenkova}.} \bibinfo{year}{2014}\natexlab{}.
\newblock \showarticletitle{Improving the Estimation of Word Importance for
  News Multi-Document Summarization}. In \bibinfo{booktitle}{\emph{Proceedings
  of the 14th Conference of the {E}uropean Chapter of the Association for
  Computational Linguistics}}. \bibinfo{publisher}{Association for
  Computational Linguistics}, \bibinfo{address}{Gothenburg, Sweden},
  \bibinfo{pages}{712--721}.
\newblock
\urldef\tempurl%
\url{https://doi.org/10.3115/v1/E14-1075}
\showDOI{\tempurl}


\bibitem[Jaidka et~al\mbox{.}(2019)]%
        {jaidka2019cl}
\bibfield{author}{\bibinfo{person}{Kokil Jaidka}, \bibinfo{person}{Michihiro
  Yasunaga}, \bibinfo{person}{Muthu~Kumar Chandrasekaran},
  \bibinfo{person}{Dragomir Radev}, {and} \bibinfo{person}{Min-Yen Kan}.}
  \bibinfo{year}{2019}\natexlab{}.
\newblock \showarticletitle{The cl-scisumm shared task 2018: Results and key
  insights}.
\newblock \bibinfo{journal}{\emph{arXiv preprint arXiv:1909.00764}}
  (\bibinfo{year}{2019}).
\newblock


\bibitem[Kedzie et~al\mbox{.}(2018)]%
        {Kedzie:2018wn}
\bibfield{author}{\bibinfo{person}{Chris Kedzie}, \bibinfo{person}{Kathleen~R
  McKeown}, {and} \bibinfo{person}{Hal Daum{\'e}~III}.}
  \bibinfo{year}{2018}\natexlab{}.
\newblock \showarticletitle{{Content Selection in Deep Learning Models of
  Summarization.}}
\newblock \bibinfo{journal}{\emph{EMNLP}} (\bibinfo{year}{2018}).
\newblock
\urldef\tempurl%
\url{https://dblp.org/rec/conf/emnlp/KedzieMD18}
\showURL{%
\tempurl}


\bibitem[Kim(2014)]%
        {Anonymous:uhC_Oukc}
\bibfield{author}{\bibinfo{person}{Yoon Kim}.} \bibinfo{year}{2014}\natexlab{}.
\newblock \showarticletitle{{Convolutional Neural Networks for Sentence
  Classification}}.
\newblock  (\bibinfo{date}{Aug.} \bibinfo{year}{2014}),
  \bibinfo{pages}{1746--1751}.
\newblock
\urldef\tempurl%
\url{http://arxiv.org/abs/1408.5882v2}
\showURL{%
\tempurl}


\bibitem[Kingma and Ba(2015)]%
        {kingma2014adam}
\bibfield{author}{\bibinfo{person}{Diederik~P Kingma} {and}
  \bibinfo{person}{Jimmy Ba}.} \bibinfo{year}{2015}\natexlab{}.
\newblock \showarticletitle{{Adam - A Method for Stochastic Optimization.}}
\newblock \bibinfo{journal}{\emph{ICLR}} (\bibinfo{year}{2015}).
\newblock
\urldef\tempurl%
\url{https://dblp.org/rec/journals/corr/KingmaB14}
\showURL{%
\tempurl}


\bibitem[Larsen and von Ins(2010)]%
        {larsen2010rate}
\bibfield{author}{\bibinfo{person}{Peder Larsen} {and} \bibinfo{person}{Markus
  von Ins}.} \bibinfo{year}{2010}\natexlab{}.
\newblock \showarticletitle{{The rate of growth in scientific publication and
  the decline in coverage provided by Science Citation Index}}.
\newblock \bibinfo{journal}{\emph{Scientometrics}} \bibinfo{volume}{84},
  \bibinfo{number}{3} (\bibinfo{year}{2010}), \bibinfo{pages}{575--603}.
\newblock


\bibitem[Luan et~al\mbox{.}(2018)]%
        {Luan:2018vn}
\bibfield{author}{\bibinfo{person}{Yi Luan}, \bibinfo{person}{Luheng He},
  \bibinfo{person}{Mari Ostendorf}, {and} \bibinfo{person}{Hannaneh
  Hajishirzi}.} \bibinfo{year}{2018}\natexlab{}.
\newblock \showarticletitle{{Multi-Task Identification of Entities, Relations,
  and Coreference for Scientific Knowledge Graph Construction}}.
\newblock \bibinfo{journal}{\emph{arXiv.org}} (\bibinfo{date}{Aug.}
  \bibinfo{year}{2018}).
\newblock
\showeprint[arxiv]{1808.09602v1}~[cs.CL]
\urldef\tempurl%
\url{http://arxiv.org/abs/1808.09602v1}
\showURL{%
\tempurl}


\bibitem[Luan et~al\mbox{.}(2017)]%
        {Luan:2017wj}
\bibfield{author}{\bibinfo{person}{Yi Luan}, \bibinfo{person}{Mari Ostendorf},
  {and} \bibinfo{person}{Hannaneh Hajishirzi}.}
  \bibinfo{year}{2017}\natexlab{}.
\newblock \showarticletitle{{Scientific Information Extraction with
  Semi-supervised Neural Tagging}}.
\newblock \bibinfo{journal}{\emph{arXiv.org}} (\bibinfo{date}{Aug.}
  \bibinfo{year}{2017}).
\newblock
\showeprint[arxiv]{1708.06075v1}~[cs.CL]
\urldef\tempurl%
\url{http://arxiv.org/abs/1708.06075v1}
\showURL{%
\tempurl}


\bibitem[Mihalcea and Tarau(2004)]%
        {W04-3252}
\bibfield{author}{\bibinfo{person}{Rada Mihalcea} {and} \bibinfo{person}{Paul
  Tarau}.} \bibinfo{year}{2004}\natexlab{}.
\newblock \showarticletitle{{TextRank: Bringing Order into Text}}. In
  \bibinfo{booktitle}{\emph{Proceedings of EMNLP 2004}}.
  \bibinfo{publisher}{Association for Computational Linguistics},
  \bibinfo{address}{Barcelona, Spain}, \bibinfo{pages}{404--411}.
\newblock
\urldef\tempurl%
\url{https://www.aclweb.org/anthology/W04-3252}
\showURL{%
\tempurl}


\bibitem[Nallapati et~al\mbox{.}(2017)]%
        {Nallapati:2017uq}
\bibfield{author}{\bibinfo{person}{Ramesh Nallapati}, \bibinfo{person}{Feifei
  Zhai}, {and} \bibinfo{person}{Bowen Zhou}.} \bibinfo{year}{2017}\natexlab{}.
\newblock \showarticletitle{{SummaRuNNer - A Recurrent Neural Network Based
  Sequence Model for Extractive Summarization of Documents.}}
\newblock \bibinfo{journal}{\emph{AAAI}} (\bibinfo{year}{2017}),
  \bibinfo{pages}{3075--3081}.
\newblock
\urldef\tempurl%
\url{https://www.scopus.com/inward/record.uri?partnerID=HzOxMe3b&scp=85030459977&origin=inward}
\showURL{%
\tempurl}


\bibitem[Narayan et~al\mbox{.}(2017)]%
        {Narayan:2017ux}
\bibfield{author}{\bibinfo{person}{Shashi Narayan}, \bibinfo{person}{Nikos
  Papasarantopoulos}, \bibinfo{person}{Shay~B Cohen}, {and}
  \bibinfo{person}{Mirella Lapata}.} \bibinfo{year}{2017}\natexlab{}.
\newblock \showarticletitle{{Neural Extractive Summarization with Side
  Information}}.
\newblock \bibinfo{journal}{\emph{arXiv.org}} (\bibinfo{date}{April}
  \bibinfo{year}{2017}).
\newblock
\showeprint[arxiv]{1704.04530v2}~[cs.CL]
\urldef\tempurl%
\url{http://arxiv.org/abs/1704.04530v2}
\showURL{%
\tempurl}


\bibitem[Ni et~al\mbox{.}(2019)]%
        {Ni:2019fb}
\bibfield{author}{\bibinfo{person}{Jianmo Ni}, \bibinfo{person}{Jiacheng Li},
  {and} \bibinfo{person}{Julian~J Mcauley}.} \bibinfo{year}{2019}\natexlab{}.
\newblock \showarticletitle{{Justifying Recommendations using Distantly-Labeled
  Reviews and Fine-Grained Aspects.}}
\newblock \bibinfo{journal}{\emph{EMNLP/IJCNLP}} (\bibinfo{year}{2019}),
  \bibinfo{pages}{188--197}.
\newblock
\urldef\tempurl%
\url{https://doi.org/10.18653/v1/D19-1018}
\showDOI{\tempurl}


\bibitem[Ni et~al\mbox{.}(2017)]%
        {Ni:2017vr}
\bibfield{author}{\bibinfo{person}{Jianmo Ni}, \bibinfo{person}{Zachary~C
  Lipton}, \bibinfo{person}{Sharad Vikram}, {and} \bibinfo{person}{Julian~J
  Mcauley}.} \bibinfo{year}{2017}\natexlab{}.
\newblock \showarticletitle{{Estimating Reactions and Recommending Products
  with Generative Models of Reviews.}}
\newblock \bibinfo{journal}{\emph{IJCNLP(1)}} (\bibinfo{year}{2017}).
\newblock
\urldef\tempurl%
\url{https://dblp.org/rec/conf/ijcnlp/NiLVM17}
\showURL{%
\tempurl}


\bibitem[Ni and McAuley(2018)]%
        {Ni:2018tg}
\bibfield{author}{\bibinfo{person}{Jianmo Ni} {and} \bibinfo{person}{Julian
  McAuley}.} \bibinfo{year}{2018}\natexlab{}.
\newblock \showarticletitle{{Personalized review generation by expanding
  phrases and attending on aspect-aware representations}}. In
  \bibinfo{booktitle}{\emph{ACL 2018 - 56th Annual Meeting of the Association
  for Computational Linguistics, Proceedings of the Conference (Long Papers)}}.
  University of California, San Diego, San Diego, United States,
  \bibinfo{pages}{706--711}.
\newblock
\showISBNx{9781948087346}
\urldef\tempurl%
\url{https://www.scopus.com/inward/record.uri?partnerID=HzOxMe3b&scp=85063144317&origin=inward}
\showURL{%
\tempurl}


\bibitem[See et~al\mbox{.}(2017)]%
        {See:2017ke}
\bibfield{author}{\bibinfo{person}{Abigail See}, \bibinfo{person}{Peter~J Liu},
  {and} \bibinfo{person}{Christopher~D Manning}.}
  \bibinfo{year}{2017}\natexlab{}.
\newblock \showarticletitle{{Get to the point: Summarization with
  pointer-generator networks}}. In \bibinfo{booktitle}{\emph{ACL 2017 - 55th
  Annual Meeting of the Association for Computational Linguistics, Proceedings
  of the Conference (Long Papers)}}. Stanford University, Palo Alto, United
  States, \bibinfo{publisher}{Association for Computational Linguistics},
  \bibinfo{address}{Stroudsburg, PA, USA}, \bibinfo{pages}{1073--1083}.
\newblock
\showISBNx{9781945626753}
\urldef\tempurl%
\url{https://doi.org/10.18653/v1/P17-1099}
\showDOI{\tempurl}


\bibitem[Stewart et~al\mbox{.}(2017)]%
        {Stewart:2017vj}
\bibfield{author}{\bibinfo{person}{Benjamin~W Stewart}, \bibinfo{person}{Andy
  Rivas}, {and} \bibinfo{person}{Luat~T Vuong}.}
  \bibinfo{year}{2017}\natexlab{}.
\newblock \showarticletitle{{Structure in scientific networks: towards
  predictions of research dynamism}}.
\newblock \bibinfo{journal}{\emph{arXiv.org}} (\bibinfo{date}{Aug.}
  \bibinfo{year}{2017}).
\newblock
\showeprint[arxiv]{1708.03850v1}~[cs.SI]
\urldef\tempurl%
\url{http://arxiv.org/abs/1708.03850v1}
\showURL{%
\tempurl}


\bibitem[Teufel and Moens(2002)]%
        {teufel-moens-2002-articles}
\bibfield{author}{\bibinfo{person}{Simone Teufel} {and} \bibinfo{person}{Marc
  Moens}.} \bibinfo{year}{2002}\natexlab{}.
\newblock \showarticletitle{Articles Summarizing Scientific Articles:
  Experiments with Relevance and Rhetorical Status}.
\newblock \bibinfo{journal}{\emph{Computational Linguistics}}
  \bibinfo{volume}{28}, \bibinfo{number}{4} (\bibinfo{year}{2002}),
  \bibinfo{pages}{409--445}.
\newblock
\urldef\tempurl%
\url{https://doi.org/10.1162/089120102762671936}
\showDOI{\tempurl}


\end{thebibliography}

\appendix

\end{document}